\pdfoutput=1

\documentclass[11pt]{article}

\usepackage[final]{acl}

\usepackage{times}
\usepackage{latexsym}

\usepackage[T1]{fontenc}

\usepackage[utf8]{inputenc}

\usepackage{microtype}

\usepackage{inconsolata}
\usepackage{url}

\usepackage{graphicx}

\usepackage{booktabs}
\usepackage{multirow}
\usepackage{enumitem} 
\usepackage[dvipsnames]{xcolor}
\usepackage[most]{tcolorbox}

\newtcolorbox{questionbox}[2][]{
    title=#2,
    colback=blue!5!white,
    colframe=blue!75!black,
    fonttitle=\small\bfseries,
    fontupper=\small,
    breakable,
    pad at break=1mm,
    #1
}

\title{AERA Chat: An Interactive Platform for\\Automated Explainable Student Answer Assessment}

\author{Jiazheng Li$^1$\thanks{Equal contribution.}\quad Artem Bobrov$^{1*}$\quad Runcong Zhao$^{1}$\quad Cesare Aloisi$^2$\quad Yulan He$^{1}$\\
  $^1$King's College London~~~$^2$AQA\\
\texttt{\{jiazheng.li, artem.bobrov, runcong.zhao\}kcl.ac.uk}\\
\texttt{caloisi@aqa.org.uk}\quad \texttt{yulan.he@kcl.ac.uk}}

\begin{document}
\maketitle
\begin{abstract}
Explainability in automated student answer scoring systems is critical for building trust and enhancing usability among educators. Yet, generating high-quality assessment rationales remains challenging due to the scarcity of annotated data and the prohibitive cost of manual verification, prompting heavy reliance on rationales produced by large language models (LLMs), which are often noisy and unreliable. To address these limitations, we present AERA Chat, an interactive visualization platform designed for automated explainable student answer assessment. AERA Chat leverages multiple LLMs to concurrently score student answers and generate explanatory rationales, offering innovative visualization features that highlight critical answer components and rationale justifications. The platform also incorporates intuitive annotation and evaluation tools, supporting educators in marking tasks and researchers in evaluating rationale quality from different models. We demonstrate the effectiveness of our platform through evaluations of multiple rationale-generation methods on several datasets, showcasing its capability for facilitating robust rationale evaluation and comparative analysis.
\end{abstract}

\section{Introduction}
Automated student answer scoring (ASAS) has become increasingly important in educational NLP applications, providing educators with efficient, consistent, and scalable methods for evaluating student responses \cite{grading_classification, helen-aes-2016, yue-aes-2017}. Traditional ASAS systems commonly leverage powerful pretrained language models as text classifiers, utilizing questions, marking rubrics, key answer elements, and student answers as input to determine scores \cite{bert_classifer_aes, xie-etal-2022-automated}. Despite their efficiency, these systems lack transparency, leading to concerns about their reliability and interpretability in practical educational contexts.

To address this limitation, prior research explored various techniques for interpreting automated scoring systems, including feature analysis \cite{tornqvist-etal-2023-exasag, bert_feature} and visual explanations such as attention-weight visualizations \cite{helen-aes-2016, yang-etal-2020-enhancing}. Nevertheless, these approaches usually demand a deep understanding of NLP methods, creating barriers for educators who are non-experts in NLP.

Recent advances in large language models (LLMs) have introduced promising approaches to explainability through natural language justifications that clearly explain scoring decisions \cite{oana-esnli, few-shot-rationalization, rationalization_survey, li-etal-2025-drift}. Such rationales enhance transparency of predictions, making automated assessments more accessible to educators and students. However, manually annotating rationale datasets for student answers remains prohibitively expensive and labor-intensive. Therefore, existing rationale-generation methods predominantly depend on potentially unreliable rationales produced by LLMs without human verification \cite{li-etal-2023-distilling, 2024_thought_tree, Li2025TwoHA}. \emph{Establishing an interactive environment for verifying, comparing, and evaluating rationales generated by multiple LLMs is essential for advancing research and ensuring trustworthy automated assessment.}

\begin{figure*}[!ht]
\centering
\includegraphics[width=\linewidth]{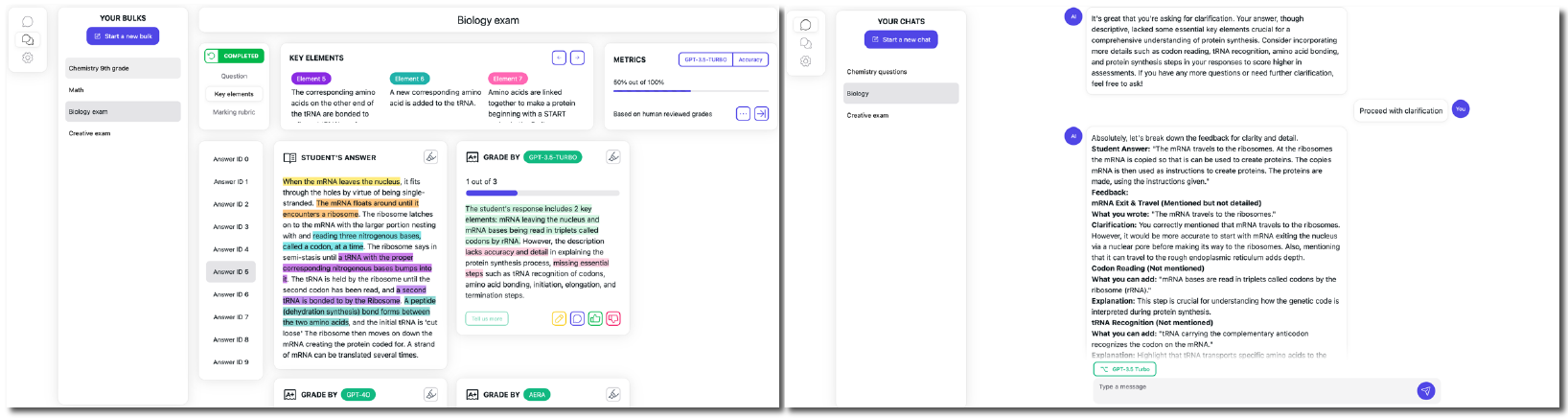}
\caption{\textbf{Overview of the two primary interfaces in the AERA Chat platform.} \emph{(1) Bulk Marking Interface (left):} Users can input question details and batch-assess student answers with automated rationale generation using multiple LLMs simultaneously. The interface highlights essential elements in student answers and rationales to facilitate easier verification. \emph{(2) Chat Interface (right):} Users can select previously assessed questions and rationales to initiate interactive discussions with LLMs, requesting more detailed explanations on assessment decisions.}
\label{fig:overview}
\vspace{-4mm}
\end{figure*}

Simultaneously, despite the increasing development of LLM-powered educational tools \cite{Wang2024LargeLM}, existing platforms mainly focus on assisting teaching, providing feedback \cite{llm_feedback_visual, TOBLER2024102531}, supporting student learning \cite{park2024thinkingassistantsllmbasedconversational, Kabir2023AnLA, 10.1007/978-3-031-64302-6_6}, or exercise generation \cite{xiao-etal-2023-evaluating, cui-sachan-2023-adaptive}, with relatively little emphasis on the assessment process itself. Therefore, a dedicated interactive platform that supports explainable, automated student answer assessment and rationale annotation has yet to be developed.

To bridge this gap, we introduce \textbf{AERA Chat}, an interactive platform explicitly designed to support automated explainable student answer scoring via LLM-generated rationales. The platform enables users, educators and researchers, to simultaneously obtain automated assessments and explanatory rationales from multiple LLMs through a unified and intuitive interface. It integrates novel visualization methods to highlight key elements in student answers and assessment rationales, significantly improving user comprehension and facilitating the verification process. Furthermore, AERA Chat supports annotation and verification processes, including rationale preference selection and direct rationale editing, streamlining human-in-the-loop data collection. The platform also provides automated scoring performance evaluation tools, helping users systematically compare and validate different LLMs’ effectiveness.

Our main contributions are:
\begin{itemize}[leftmargin=*,noitemsep,topsep=0pt]
\item We developed an innovative interactive platform for \textbf{automated explainable student answer scoring} using multiple LLMs, deployable via public APIs or private instances.
\item We introduced intuitive visualization techniques to clearly \textbf{highlight critical components in student answers and rationales}, enhancing readability and verification efficiency.
\item Our platform supports streamlined rationale annotation and verification through \textbf{built-in preference selection and direct annotation tools}, simplifying rationale-quality evaluations and annotation processes.
\item We evaluated the assessment capabilities of three different rationale-generation models using four publicly available datasets and two proprietary datasets, demonstrating the effectiveness and versatility of AERA Chat.
\end{itemize}

\noindent A demonstration video of our system is available at \url{https://youtu.be/1GZhTpNvMw4}. An online demonstration of the system can also be accessed at \url{https://aerachat.sites.er.kcl.ac.uk/}.

\section{Overview of AERA Chat}
\begin{figure*}[ht]
\centering
\includegraphics[width=\linewidth]{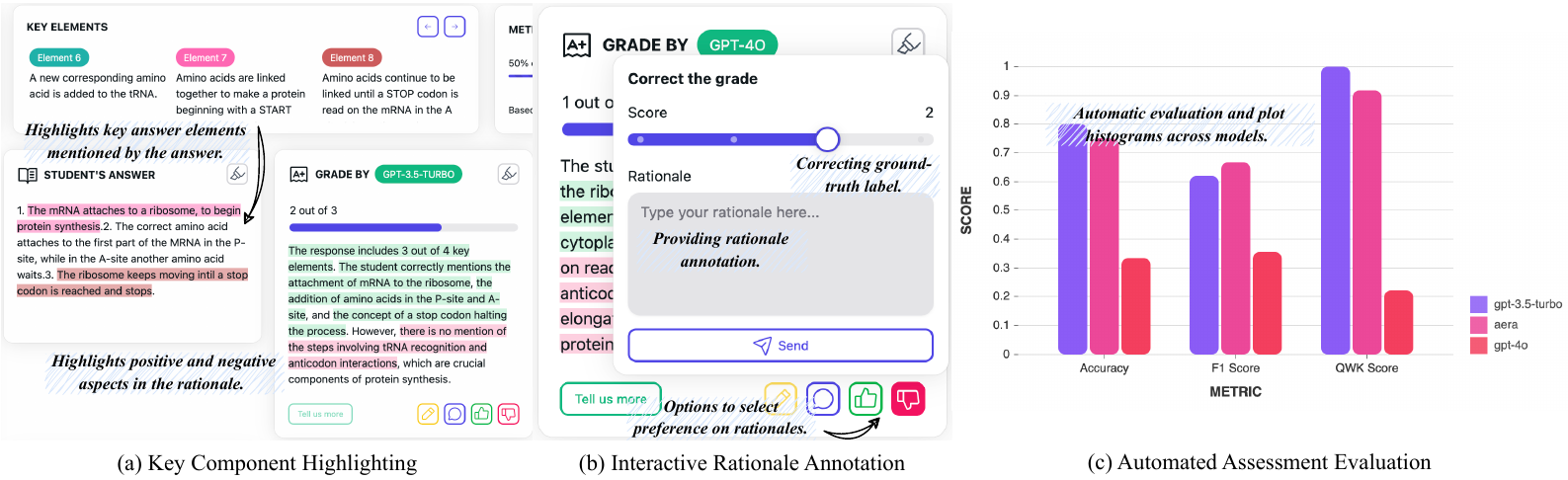}
\caption{\textbf{Primary functionalities available in the Bulk Marking Interface.}}
\label{fig:feature_functions}
\vspace{-4mm}
\end{figure*}

\subsection{Bulk Marking Interface} \label{sec:bulk_marking}
As depicted on the left-hand side of Figure \ref{fig:overview}, the bulk marking interface enables users to configure a question along with a collection of student responses to be evaluated. Once these inputs are submitted, our backend concurrently processes the student responses and sends queries to all selected LLMs to obtain both a scoring decision and an accompanying rationale that explains the assessment. After processing is complete, our platform can emphasize the key components within student responses or rationales, drawing attention to the most pertinent contextual elements. Additionally, the platform automatically computes the assessment performance of the LLMs and presents the results through histograms. The interactive platform further facilitates the annotation or verification of the generated rationales.

\paragraph{Bulk Initialization} In the bulk initialization phase, users define the question, specify key answer elements (the critical key phrases required for an accurate response), and establish a point-based marking rubric (criteria that allocate marks based on these key elements). The system then automatically aggregates the question details into a template prompt, referred to as $Q_{\textnormal{info}}$, which is subsequently provided to the LLMs for evaluation. Users also have the option to upload a file containing a batch of student responses to be assessed, which may include corresponding ground-truth marks. We denote the collection of uploaded student responses and, if available, their associated ground-truth marks as $D = {(x_i, y_i)}_{i=1}^N$, where $N$ represents the total number of student responses for the given question.

\paragraph{Automated Scoring and Rationale Generation}
Once the question details and student responses are uploaded to the system, our platform proceeds to automatically evaluate the student responses and generate corresponding rationales using the selected LLMs. For illustration purposes, we employed two OpenAI models: \texttt{GPT-3.5-turbo} and \texttt{GPT-4o}, in addition to our in-house rationale generation model—the AERA model \cite{li-etal-2023-distilling}. Our system is designed to be easily extended to incorporate other models for the purpose of student response assessment and rationale generation.

To meet the requirement of simultaneously producing scores and free-form textual rationales, we instruct the LLM $\texttt{LLM}_{\theta}$ to output its results in a JSON format \cite{overprompt}. This process is represented as: $(\hat{y}_{i}, \hat{r}_{i}) = \texttt{LLM}_{\theta}(x_i, Q_{\textnormal{info}})$, where $\hat{y}_{i}$ denotes the predicted mark for the student response and $\hat{r}_{i}$ is a textual rationale that explains the marking decision.

As shown on the left-hand side of Figure \ref{fig:overview}, once the assessment is finalized, the platform displays the rationales generated by each LLM in parallel in a card-based view. The score assigned by each LLM appears at the top of its respective card, enabling users to effortlessly compare the scoring decisions across different LLMs.

\paragraph{Key Component Highlighting}
Ensuring that the assessment rationales faithfully represent the student’s answer is a challenging task that demands a deep comprehension of the context. Our platform is designed to improve the user’s reading experience by clearly marking the relevant portions of both student responses and assessment rationales using high-contrast colours. In contrast to the visualization methods described in previous studies \cite{helen-aes-2016, yang-etal-2020-enhancing}, our highlighting feature functions as a stand-alone semantic comparison tool that supports users in understanding the reasoning behind the LLM’s decisions.

As illustrated in Figure \ref{fig:feature_functions} (a), users have the option to either display the key answer elements found within the student responses in a uniform colour or to differentiate between positive aspects (indicating reasons for awarding points) and negative aspects (indicating reasons for point deductions) within the LLM-generated rationales. This functionality is implemented by automatically querying GPT-4o for word-level tagging, with our backend subsequently processing these annotations and applying contrasting colours to highlight the context.

\paragraph{Rationale Verification Tool}
Given that no automated metric currently exists for evaluating free-text rationale quality, most assessments of rationale quality are conducted manually. To support the evaluation and annotation of rationales, we have incorporated three distinct functionalities, as demonstrated in Figure \ref{fig:feature_functions} (b):

\noindent \textbf{(1) Label correction}: As highlighted in \cite{li-etal-2023-distilling}, some ground-truth labels in the publicly available ASAS dataset \cite{asap-aes} may be inaccurate. Consequently, we provide users with the option to amend ground-truth labels. This feature leverages the multi-LLM assessment capability; when multiple LLMs concur on a specific score, users are prompted to review and potentially correct the original label if it appears to be erroneous.

\noindent \textbf{(2) Rationale preference selection}: Assessing the quality of rationales is inherently challenging due to the absence of standardized evaluation metrics. Existing research has often framed this qualitative evaluation as a binary preference task \cite{li-etal-2023-distilling, 2024_thought_tree}, where the rationale that is both factually accurate and more detailed is favoured. As a result, our platform includes options for users to indicate whether they “prefer” or “do not prefer” a given rationale. These choices are automatically logged in the system’s database, enabling researchers to compile preference data that can inform the latest reinforcement learning from human feedback techniques \cite{dpo,rlhf} for training rationale generation LLMs.

\noindent \textbf{(3) Direct annotation}: In cases where none of the candidate rationales produced by the LLMs is deemed correct, our platform offers users the ability to submit their own annotated assessment rationales. This annotated data can then be used for supervised fine-tuning, further enhancing the model’s performance.

\paragraph{Assessment Performance Evaluation}
If users have provided ground-truth labels for the student responses via our interface, as depicted in Figure \ref{fig:feature_functions} (c), our platform will automatically assess the performance of the evaluation decisions across the chosen LLMs. This performance is illustrated using a histogram that displays metrics such as Accuracy, macro F1 Score, and QWK (Quadratic Weighted Kappa) score. We utilize the metric implementations available in the Sci-kit Learn package\footnote{\url{https://scikit-learn.org/}}.

\subsection{Chat Interface} \label{sec:chat_marking}
To capitalize on the rich conversational abilities of LLMs and to utilize their ability to provide explainable scoring of student responses, our platform includes a chat interface, illustrated on the right-hand side of Figure \ref{fig:overview}. This interface enables users to interact with LLMs by incorporating both the question details and the rationales generated from the bulk marking system.

Educators can leverage these chat functions to request more in-depth explanations when the assessment rationales appear ambiguous. In contrast, researchers have the opportunity to engage with LLMs to reflect upon any incorrect assessment rationales and generate revised evaluation decisions. The Chat Interface offers a variety of LLM options; should a user find the current LLM’s response unsatisfactory, they can switch to a more advanced model or a task-specific model. Additionally, all chat interactions are automatically logged in our system’s database, providing researchers with a rich source of training data to further refine the rationale generation capabilities of LLMs.

\section{Implementation Details}
AERA Chat is built on a microservices architecture that unifies multiple LLMs through a single web interface. We utilize Docker to modularize the services, which not only improves modularity but also enhances the system’s scalability and maintainability. This strategy facilitates flexible deployment of AERA Chat, whether on a single server or distributed across multiple servers.

\begin{figure}[ht]
\centering
\includegraphics[width=\linewidth]{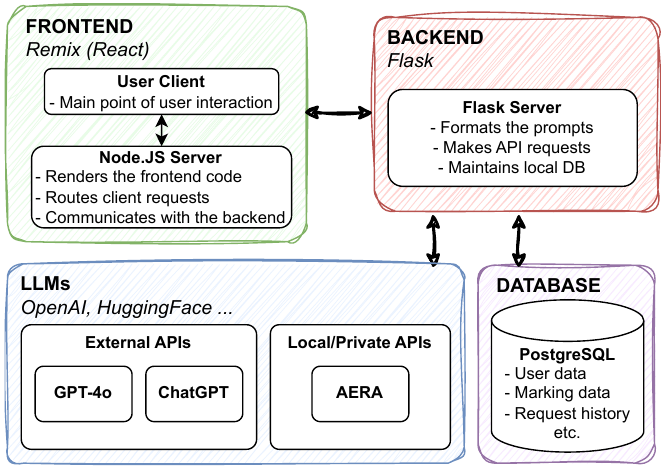}
\vspace{-2mm}
\caption{\textbf{System Architecture of AERA Chat.}}
\vspace{-2mm}
\label{fig:architecture}
\end{figure}

As illustrated in Figure \ref{fig:architecture}, the system architecture of AERA Chat is comprised of four primary components: the Frontend, the Aggregation Backend Layer, the LLM Services, and the Database.

\paragraph{Frontend} \label{frontend}
The frontend delivers a responsive web interface designed specifically for educators and researchers to interact with the system. It is engineered with usability as a priority and offers features such as login, registration, chat functionality, bulk assessments, and comprehensive feedback options. Developed using the Remix framework based on React, the frontend leverages isomorphic TypeScript primarily for server-side rendering and employs a nested routing strategy. This design approach effectively shifts a significant portion of processing from the client side to the server side, ensuring high performance and a modern user experience.

\begin{table*}[!ht]
\centering
\resizebox{\linewidth}{!}{%
\begin{tabular}{@{}l|ccc|ccc|ccc|ccc|ccc|ccc@{}}
\toprule
\multirow{2}{*}{\textbf{Model}} & \multicolumn{3}{c|}{\textbf{ASAP \#1} (Science)} & \multicolumn{3}{c|}{\textbf{ASAP \#2} (Science)} & \multicolumn{3}{c|}{\textbf{ASAP \#5} (Biology)} & \multicolumn{3}{c|}{\textbf{ASAP \#6} (Biology)} & \multicolumn{3}{c|}{\textbf{Prop 1} (Biology)} & \multicolumn{3}{c}{\textbf{Prop 2} (Biology)} \\
\cmidrule(l){2-19}
 & Acc & F1 & QWK & Acc & F1 & QWK & Acc & F1 & QWK & Acc & F1 & QWK & Acc & F1 & QWK & Acc & F1 & QWK \\
\midrule
\texttt{gpt-4o}      & 0.641 & 0.635 & \underline{0.792} & \underline{0.615} & \underline{0.608} & \underline{0.759} & 0.825 & \underline{0.805} & 0.695 & 0.793 & \textbf{0.817} & 0.721 & \underline{0.606} & \underline{0.602}  & \underline{0.802} & 0.452 & 0.458 & 0.761 \\
\texttt{o3-mini}     & 0.579 & 0.570 & 0.751 & 0.582 & 0.568 & 0.751 & 0.726 & 0.754 & 0.664 & 0.713 & 0.765 & 0.561 & 0.597 & 0.591 & 0.761 & \underline{0.500} & \underline{0.481} & \textbf{0.783} \\
\midrule
\texttt{Llama-3-70B} & 0.594 & 0.597 & 0.720 & 0.467 & 0.446 & 0.609 & 0.813 & \textbf{0.813} & 0.713 & 0.773 & \underline{0.811} & 0.683 & 0.602 & 0.601 & 0.780 & 0.398 & 0.384 & 0.569 \\
\texttt{Qwen-32B}    & 0.323 & 0.307 & 0.189 & 0.392 & 0.403 & 0.232 & 0.706 & 0.717 & 0.413 & 0.775 & 0.781 & 0.472 & 0.388 & 0.376 & 0.200 & 0.339 & 0.332 & 0.360 \\
\texttt{Qwen2.5-72B} & 0.186 & 0.058 & -0.001 & 0.160 & 0.044 & 0.000 & 0.773 & 0.674 & 0.000 & 0.831 & 0.758 & 0.020 & 0.225 & 0.082 & 0.000 & 0.213 & 0.103 & 0.004 \\
\midrule
\texttt{AERA}        & \underline{0.654} & \underline{0.658} & 0.765 & 0.547 & 0.550 & 0.734 & \underline{0.861} & 0.567 & \underline{0.818} & \textbf{0.888} & 0.579 & \underline{0.802} & 0.406 & 0.384 & 0.714 & 0.255 & 0.102 & 0.002 \\
\texttt{Thought Tree}& \textbf{0.774} & \textbf{0.781} & \textbf{0.875} & \textbf{0.653} & \textbf{0.676} & \textbf{0.779} & \textbf{0.878} & 0.633 & \textbf{0.845} & \underline{0.872} & 0.601 & \textbf{0.825} & \textbf{0.621} & \textbf{0.628} & \textbf{0.844} & \textbf{0.587} & \textbf{0.592} & \underline{0.780} \\
\bottomrule
\end{tabular}%
}
\vspace{-2mm}
\caption{\textbf{Evaluation of Assessment Performance.} The best score is shown in \textbf{bold} and the second-best is \underline{underlined}.}
\vspace{-5mm}
\label{tab:model_comparison_corrected}
\end{table*}

\paragraph{Aggregation Backend Layer} \label{backend}
Serving as the core of the AERA Chat platform, the backend layer integrates all other components. It functions via a REST API, managing HTTP requests without maintaining persistent connection states. Implemented with the Flask framework, this layer also incorporates WebSocket technology to establish persistent connections. This integration enables clients to receive real-time updates directly from the server, eliminating the need for constant refresh requests.

\paragraph{LLM Services} \label{llm_service}
Assessing student responses is a critical task in educational environments. Accordingly, our system offers users the choice between using publicly available API-based LLMs, such as GPT-4 \cite{gpt4} or Gemini \cite{gemini}, and deploying privately trained, custom LLMs to achieve enhanced performance on specific questions or to ensure better data privacy. %

\paragraph{Database} \label{database}
The database layer is powered by a PostgreSQL relational database, which is used to manage a wide range of data, including user profiles, assessment records (both for real-time tracking and background processing), and chat histories. It interfaces with the backend layer via SQLAlchemy, an object-relational mapping tool, to ensure both robust security and high speed. %

\section{Universal Evaluation}
We utilized AERA Chat's built-in tools to conduct a comprehensive evaluation of several LLMs on automated student scoring.

\paragraph{Datasets} We used four subsets from the widely-recognized ASAP-AES dataset \cite{asap-aes} and two proprietary (Prop) Biology datasets. These datasets cover Science and Biology subjects with varying score ranges. Table \ref{tab:data_statistic} summarizes the test set statistics.

\begin{table}[!h]
\centering
\resizebox{\linewidth}{!}{
\begin{tabular}{lcccccc}
\toprule
\textbf{Datasets} & ASAP \#1 & ASAP \#2 & ASAP \#5 & ASAP \#6 & Prop 1 & Prop 1 \\
\midrule
Test & 554 & 426 & 598 & 599 & 254 & 196 \\
Score Range & 0-3 & 0-3 & 0-3 & 0-3 & 0-4 & 0-3 \\
\bottomrule
\end{tabular}}
\caption{\textbf{Test set statistics.}}
\label{tab:data_statistic}
\end{table}

\paragraph{Models} We evaluated a diverse set of models: \textbf{Proprietary APIs}: gpt-4o-2024-11-20 and o3-mini-2025-01-31. \textbf{Open-Source Models}: Llama-3.3-70B, Qwen-QwQ-32B, and Qwen2.5-72B. \textbf{Our Local Specialized Models}: AERA \cite{li-etal-2023-distilling} and Thought Tree \cite{2024_thought_tree}, which are specifically designed for rationale generation.

\subsection{Evaluation of Assessment Performance}
The assessment performance results, automatically computed by AERA Chat, are presented in Table 1. Our analysis reveals several key insights:

\paragraph{Specialized Models Excel} The Thought Tree model consistently emerges as the top performer across nearly all datasets, achieving the highest QWK scores, often by a significant margin. This highlights the benefit of specialized architectures for this task. The AERA model also demonstrates strong, competitive performance, particularly on the ASAP datasets it was trained on, underscoring the value of in-domain fine-tuning.

\paragraph{Proprietary vs. Open-Source LLMs} Among the general-purpose models, gpt-4o delivers solid and reliable performance, generally outperforming the open-source alternatives. The o3-mini model also shows respectable results, suggesting that smaller, efficient proprietary models can be effective. In contrast, the performance of the open-source models is highly variable. While Llama-3.3-70B is competitive on some datasets, the Qwen models, particularly Qwen2.5-72B, struggle significantly, with QWK scores near or below zero, indicating their predictions are no better than random chance for this task.

\paragraph{Subject-Specific Challenges} Performance tends to be lower on the Science datasets (\#1 and \#2) compared to the Biology datasets. This may be attributed to the more complex reasoning required for the science questions, which involve analyzing experimental designs. This suggests that future model development should focus on improving performance on questions requiring deeper contextual understanding.

\subsection{Human Evaluation of Rationale Quality}
To complement our quantitative analysis, we conducted a rigorous human evaluation of the generated rationales using AERA Chat's annotation interface. This evaluation focused on the top-performing models: Thought Tree, GPT-4o, and Llama-3-70B.

Following the same annotation rubric as in prior work \cite{2024_thought_tree}, two expert graders independently inspected 30 randomly-drawn student responses from each of the six test sets. Each rationale was judged on: \textbf{Correctness}: \emph{Does the explanation faithfully justify the score in line with the official rubric?} \textbf{Preference}: \emph{In a blind pairwise comparison between models for the same student answer, which rationale is more helpful and appropriate as feedback?}\footnote{Inter-rater agreement (Cohen's $\kappa$) was 0.82 for correctness and 0.76 for preference, indicating substantial consensus.} 

\begin{table}[h]
\centering
\resizebox{0.9\linewidth}{!}{%
\begin{tabular}{@{}lccc@{}}
\toprule
\textbf{Metric}      & \textbf{GPT-4o} & \textbf{Llama 3 70B} & \textbf{Thought Tree} \\ \midrule
Correctness (\%)     & 68              & 44                   & \textbf{76}           \\
Preference Win Rate (\%) & 28              & 22                   & \textbf{50}           \\ \bottomrule
\end{tabular}%
}
\caption{\textbf{Human evaluation of rationale quality}.}
\label{tab:human_eval_main}
\end{table}

As shown in Table \ref{tab:human_eval_main}. Thought Tree approach delivered the most accurate and most preferred rationales. Annotators praised its habit of citing rubric points verbatim and linking them directly to concrete excerpts from the student's answer. While GPT-4o’s explanations were usually on target, reviewers noted occasional omissions in evaluation details (e.g., missing a required rubric element). Its writing style is concise; some annotators preferred the more granular, point-by-point feedback of other models, which hurt its preference share. Llama-3.3-70B produced fluent text but struggled with identifying partial-credit elements. This led to inconsistencies, such as awarding full marks while the rationale only discussed half of the required rubric points.

\subsection{User Study for AERA Chat}
To validate the usability and effectiveness of AERA Chat, we conducted a user study with 16 participants, comprised of 8 educators and 8 NLP researchers. Participants were tasked with using the platform to assess and annotate a set of 15 student answers. Following the task, they completed a detailed questionnaire (see Appendix \ref{sec:questionnaire}). 

\begin{table}[h!]
\centering
\resizebox{\linewidth}{!}{%
\begin{tabular}{@{}lc@{}}
\toprule
\textbf{Questionnaire Item} & \textbf{Average Score} \\ \midrule
\multicolumn{2}{l}{\textit{\textbf{Usability \& Interface Design}}} \\
Q1: Ease of Navigation & 4.4 / 5 \\
Q2: Intuitive Layout for Comparison & 4.8 / 5 \\ \midrule
\multicolumn{2}{l}{\textit{\textbf{Assessment \& Rationale Quality}}} \\
Q4: Accuracy of Scores & 3.9 / 5 \\
Q5: Quality of Rationales & 4.6 / 5 \\
Q6: System Speed & 4.8 / 5 \\ \midrule
\multicolumn{2}{l}{\textit{\textbf{Explainability \& Verification Tools}}} \\
Q7: Helpfulness of Highlighting & 5.0 / 5 \\
Q8: Ease of Annotation Tools & 4.3 / 5 \\ \midrule
\multicolumn{2}{l}{\textit{\textbf{Research \& Comparative Utility}}} \\
Q10: Effectiveness for Comparing LLMs & 4.5 / 5 \\
Q11: Usefulness of Performance Metrics & 4.4 / 5 \\
Q12: Support for Data Collection & 4.1 / 5 \\ \bottomrule
\end{tabular}%
}
\caption{\textbf{Average user ratings from the questionnaire.}}
\label{tab:user_study_results}
\end{table}

The platform's user interface was highly rated for its ease of use. The side-by-side view for comparing LLM-generated scores and rationales was noted as intuitive. Users reported that the quality of the AI-generated rationales was high, and they praised the system's responsiveness and speed.

The standout feature of the platform was the Key Component Highlighting. Every participant described this feature as ``Very Helpful'' for understanding and verifying the LLM's decisions at a glance. This powerful explainability tool translated directly into user confidence; 88\% of participants reported that the platform ``significantly'' increased their trust in the automated assessment compared to systems that only output a score.

The study confirmed the platform's value for its two target audiences. For Educators, the primary advantages were ``Speed and Efficiency'' and ``Improved Assessment Consistency''. For Researchers, the platform excelled as a tool for ``Comparing Different Models'' and ``Facilitating Data Annotation''. Compared to previous methods, AERA Chat was seen as a major improvement. When asked about future adoption, 94\% of participants confirmed they would use the platform again. 

\section{Conclusion}
In this paper, we introduced AERA Chat, an interactive platform designed to make automated student answer assessment transparent and trustworthy. Our comprehensive evaluations, conducted within the platform, reveal that specialized models deliver superior scoring and rationale quality over general-purpose LLMs, while key explainability features, like our key component highlighting, directly translate into significant gains in user trust. By integrating assessment, explanation, and collaborative annotation, AERA Chat provides a dual-use solution, acting as both \emph{an efficiency tool for educators} and \emph{a robust research workbench}, that marks a significant step towards developing more accountable and effective AI for education.

\section*{Acknowledgements}

This work was supported in part by the UK Engineering and Physical Sciences Research Council through an Impact Accelerator Account at King's College London (grant no. EP/X525571/1), a Turing AI Fellowship (grant no. EP/V020579/1, EP/V020579/2),  and a Prosperity Partnership project with AQA (UKRI566). JL is funded by a PhD scholarship provided by AQA.

\bibliography{custom}

\appendix

\section{Further Evaluation and Implementation Details}
\label{sec:appendix}

\subsection{Experimental Setup}

\paragraph{Datasets}
The evaluation was conducted on six distinct datasets. Four are public subsets of the ASAP-AES dataset \cite{asap-aes}, and two are proprietary datasets provided by our research partner.

\textbf{ASAP Datasets (\#1, \#2, \#5, \#6):} These datasets feature short-answer questions from Science and Biology domains. We used the official train/test splits. The tasks range from interpreting experimental setups (Science) to demonstrating biological knowledge (Biology).

\textbf{Proprietary Datasets (Prop 1, Prop 2):} These datasets consist of student responses to GCSE-level Biology questions. They represent a more contemporary and challenging assessment scenario.

\paragraph{Models and Parameters}
The models used in our evaluation were accessed via their official APIs or hosted locally.

\textbf{API-based Models:} We used gpt-4o (via gpt-4o-2024-05-13) and o3-mini through the OpenAI API.
\textbf{Locally Hosted Models:} Open-source models were downloaded from Hugging Face and run using the Transformers library. These included Llama-3.3-70B (meta-llama/Meta-Llama-3.3-70B-Instruct), the Qwen series, the AERA model (jiazhengli/long-t5-tglobal-large-AERA), and Thought Tree (jiazhengli/Mixtral-8x7B-Instruct-v0.1-QLoRA-Assessment-Rationale-dpo).
For all LLM-based evaluations, a consistent generation configuration was used: a temperature of `0.7' and a maximum output token limit of `512'.

\subsection{Prompt Template}
All models were prompted using a structured template designed to provide all necessary context for the assessment task. The template instructs the model to return a JSON object containing the score and the rationale.

\begin{questionbox}[breakable]{LLM Assessment Prompt Template}
You are an expert AI assistant tasked with grading a student's answer.
Please assess the following student response based on the provided question, marking rubric, and key answer elements.

Your output MUST be a single JSON object with two keys: "score" (an integer) and "rationale" (a string explaining your reasoning).

\rule{\linewidth}{0.4pt}

\textbf{Question:}
[Question text goes here]

\textbf{Key Answer Elements:}
[List of key phrases or concepts required for a perfect score goes here]

\textbf{Marking Rubric:}
[Full marking rubric, describing how points are awarded, goes here]

\rule{\linewidth}{0.4pt}

\textbf{Student Answer:}
[The student's raw answer text goes here]

\rule{\linewidth}{0.4pt}

\textbf{Your JSON Output:}
\end{questionbox}

\subsection{Human Evaluation Protocol Details}

\paragraph{Annotator Instructions}
Two expert graders, both with experience in educational assessment, were recruited. They were given access to the AERA Chat interface and provided with the following instructions for the two evaluation tasks.

\textbf{Task 1: Correctness Evaluation}
For each rationale presented:
\begin{enumerate}
    \item Read the student answer, the official question, and the marking rubric carefully.
    \item Read the rationale generated by the model.
    \item Assess two criteria:
        \begin{itemize}
            \item \textit{Factual Alignment:} Does the rationale correctly identify which key elements the student did or did not mention?
            \item \textit{Score Justification:} Does the final score assigned by the model logically follow from the justification given in its rationale?
        \end{itemize}
    \item Mark the rationale as \textbf{"Correct"} only if both criteria are fully met. Otherwise, mark it as \textbf{"Incorrect"}.
\end{enumerate}

\textbf{Task 2: Preference Evaluation}
For each student answer, you will be shown the rationales from three different models in a random, blind order.
\begin{enumerate}
    \item Review all rationales for the same student answer.
    \item Select the single rationale that you believe provides the most helpful and clear feedback for a student.
    \item Your preference should be based on clarity, tone, accuracy of detail, and educational value. You are selecting the explanation you would be most willing to share with the student.
\end{enumerate}

\section*{User Questionnaire}
\label{sec:questionnaire}

To evaluate the usability and effectiveness of \textbf{AERA Chat}, we designed a structured questionnaire for our target users (educators and researchers). The survey is organized into five key dimensions: Usability and Interface Design, Automated Assessment and Rationale Quality, Explainability and Verification Tools, Comparative Analysis and Research Utility, and Overall Experience and Impact. The full questionnaire is detailed below.

\begin{questionbox}{Section 1: Usability and Interface Design}
    \textbf{Q1.} How would you rate the ease of navigating the AERA Chat platform, including both the Bulk Marking and Chat interfaces? (1 = Very Difficult, 5 = Very Easy)
    
    \vspace{1.5em}
    \textbf{Q2.} How intuitive is the layout for comparing scores and rationales from multiple LLMs side-by-side? (1 = Not Intuitive, 5 = Very Intuitive)
    
    \vspace{1.5em}
    \textbf{Q3.} Is the purpose of the two main interfaces (Bulk Marking vs. Chat) clear and distinct? \\
    - Yes \\
    - No \\
    - Somewhat
\end{questionbox}

\begin{questionbox}{Section 2: Automated Assessment and Rationale Quality}
    \textbf{Q4.} How would you rate the accuracy of the scores generated by the different models? (1 = Very Inaccurate, 5 = Very Accurate)
    
    \vspace{1.5em}
    \textbf{Q5.} Please assess the overall quality of the AI-generated rationales in terms of clarity, correctness, and justification for the score. (1 = Poor, 5 = Excellent)
    
    \vspace{1.5em}
    \textbf{Q6.} How responsive and fast is the system when generating assessments for a batch of student answers? (1 = Very Slow, 5 = Very Fast)
\end{questionbox}

\begin{questionbox}{Section 3: Explainability and Verification Tools}
    \textbf{Q7.} How helpful is the "Key Component Highlighting" feature for understanding and verifying the LLMs' assessment decisions? (1 = Not Helpful, 5 = Very Helpful)
    
    \vspace{1.5em}
    \textbf{Q8.} How easy is it to use the annotation tools (e.g., correcting ground-truth labels, selecting rationale preference, editing rationales)? (1 = Very Difficult, 5 = Very Easy)
    
    \vspace{1.5em}
    \textbf{Q9.} Does AERA Chat's explainability (rationales + highlighting) increase your trust in the automated assessment process compared to a system that only outputs a score? \\
    - Yes, significantly \\
    - Yes, somewhat \\
    - No difference \\
    - No, it reduces my trust
\end{questionbox}

\begin{questionbox}{Section 4: Comparative Analysis and Research Utility}
    \textbf{Q10.} How effective is the platform for comparing the performance and behaviour of different LLMs for student answer assessment? (1 = Not Effective, 5 = Very Effective)
    
    \vspace{1.5em}
    \textbf{Q11.} How useful are the automatically calculated performance metrics (Accuracy, F1, QWK) for your evaluation needs? (1 = Not Useful, 5 = Very Useful)
    
    \vspace{1.5em}
    \textbf{Q12.} Does the platform effectively support the collection of human feedback data (e.g., rationale preferences, corrections) for research purposes? (1 = Not at all, 5 = Extremely well)
\end{questionbox}

\begin{questionbox}{Section 5: Overall Experience \& Impact}
    \textbf{Q13.} Before using AERA Chat, how did you typically perform tasks like scoring student answers or evaluating AI-generated explanations? (Select all that apply) \\
    - Manually graded and wrote justifications myself \\
    - Used general-purpose chatbots (e.g., ChatGPT) \\
    - Used other specialized assessment tools \\
    - Did not perform this type of task
    
    \vspace{1.5em}
    \textbf{Q14.} Compared to your previous method(s), what are the primary advantages of using AERA Chat? (Select up to three) \\
    - Speed and Efficiency \\
    - Explainability and Transparency \\
    - Ease of Comparing Different Models \\
    - Facilitates Data Annotation/Collection \\
    - Improved Assessment Accuracy/Consistency \\
    - Other: \underline{\hspace{3.5cm}}
    
    \vspace{1.5em}
    \textbf{Q15.} For future tasks involving student answer assessment or rationale evaluation, how likely are you to use AERA Chat again? \\
    - I would prefer to use AERA Chat \\
    - I would combine AERA Chat with my previous methods \\
    - I would revert to my previous methods \\
    - I am unsure
\end{questionbox}

\end{document}